\documentclass[runningheads]{llncs}
\usepackage{graphicx}
\begin{document}
% Deep Ensembled Netwok for Visual
\title{Improving Landmark Recognition using Saliency detection and  Feature classification}
% \thanks{Supported by organization x.}
%%\orcidID{0000-1111-2222-3333
% \author{Anonymous Author(s)***}
\author{Akash Kumar\inst{*} \and
Sagnik Bhowmick\inst{*} \and 
% \thanks{Equal Contribution} \and
N. Jayanthi \and
S. Indu }
\authorrunning{Akash \& Sagnik et al.}
\institute{Delhi Technological University, New Delhi \\
\email{\{akash\_bt2k15, sagnikbhomic\_2k15ec133, njayanthi, s.indu\}@dtu.ac.in}\\
}
\maketitle              % typeset the header of the contribution
\begin{abstract}
Image Landmark Recognition has been one of the most sought-after classification challenges in the field of vision and perception. After so many years of generic classification of buildings and monuments from images, people are now focussing upon fine-grained problems - recognizing the category of each building or monument. We proposed an ensemble network for the purpose of classification of Indian Landmark Images. To this end, our method gives robust classification by ensembling the predictions from Graph-Based Visual Saliency (GBVS) network alongwith supervised feature-based classification algorithms such as kNN and Random Forest. The final architecture is an adaptive learning of all the mentioned networks.  The proposed network produces a reliable score to eliminate false category cases. Evaluation of our model was done on a new dataset, which involves challenges such as landmark clutter, variable scaling, partial occlusion, etc. 

\end{abstract}

\footnote{* Equal Contribution}

\section{Introduction}
United Nation Educational, Scientific and Cultural Organization (UNESCO) World Heritage Center recognizes over 1500 monuments and landmarks as World heritage sites. Apart from this, there are over 10K monuments and landmarks spread over the globe which serves as a local tourist attraction and have a huge contribution in the history and culture of the location. However, it is impossible for humans to individually recognize and classify all monuments according to history and architecture. Technology like Computer Vision and Deep Learning plays a pivotal role to overcome this challenge. 

Many Convolutional Neural Network(CNN) based deep learning frameworks shows to be handy in such a scenario, where classes have different features. Every landmark architecture style has distinguishable features from other forms of architecture. These features play a pivotal role in the recognition of such landmark architectures. India, one of the most diverse country in the world, is a house to varied architectures. We propose a framework to classify these landmarks based on the era they were constructed.  These varied architectural features make classiﬁcation of Indian monuments a dreadful task. Moreover, these historic buildings are useful references for architects designing contemporary architecture, thus information about the architectural styles of these monuments seems necessary.

In this paper, we employ CNN to address the problem of Landmark Recognition. Our main contributions are:
\begin{enumerate}
    \item We proposed an end-to-end architecture to classify Indian monuments in the image. Experiments show that our model surpass the existing baseline on the dataset.
    \item We employ convolutional architectures to learn the intra-class variations between different landmarks. The final the averageprediction is 
    ensemble of three networks consisting of salient regions detection, kNN and Random Forest supervised classification algorithms.
\end{enumerate} 

\section{Related Work}

There are several recent papers to address the problem of Landmark Recognition \cite{sift}, \cite{ref_1}, \cite{ref_2}, most of them are based on deep learning except \cite{ref_7}, which classify landmarks using visual features such as HoG \cite{hog}, SIFT \cite{sift} and SURF \cite{surf}. While landmark recognition can be considered as descriptor matching, our work relates to some \cite{ref_1} in that we learn to employ a visual saliency algorithm to focus on the most noticeable region and extract those features to classify them. 

Landmark Recognition using CNN presents a competitive research as there is so much little intra-class variations \cite{ref_url1}. \cite{ref_2} employed a multi-scale feature embedding to generate condition viewpoint invariant features for specific place recognition. \cite{ref_10} uses local binary patterns and Gray-level co-occurrence matrix to match the pairs using pixel-wise information. \cite{ref_11} devised an architecture using visual descriptors and Bag-of-Words for Image-based Monument Classification. \cite{ref_1} uses AlexNet to extract features and classify landmarks using supervised feature learning. Many works has been done on specific place recognition but the area of using fine-grained features to recognize Indian Landmarks has not been explored yet. 

\section{Problem Formulation}
Landmark Recognition in Indian scenario is very different from the European and American counter-part, due to its extreme varieties within each region and diversified architecture. Approaches like bag of words, HoG and SIFT are constrained to database size. Other approaches based on deep learning framework \cite{ref_8} face challenges in identifying diverse image in same class (refer fig.1). Among all these methods, we need a more robust and dynamic framework that can learn these intra-class variations. Hence, our architecture focusses on these explicit and implicit features of the images.

\begin{figure}
\includegraphics[width=\textwidth]{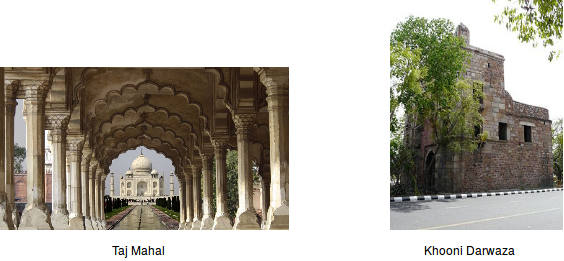}
\caption{Same class had varied architecture style} \label{fig1}
\end{figure}

\section{Dataset}
The manually collected Indian monument dataset consist of monument images majorly of 4 classes based on architecture types, i.e. Buddhist, Dravidian, Kalinga and Mughal architectures.    

\begin{table}[]
    \centering
    \caption{Categorical distribution of data among 4 classes}\label{tab1}
    \begin{tabular}{|l|l|l|l|}
    \hline
    {\large Class Label} & {\large Train Set} & {\large Validation Set} & {\large Test Set}\\
    \hline
    {\bfseries I. Buddhist} &  647 & 81 & 81\\ \hline 
    {\bfseries II. Dravidian} & 657 & 83 & 82\\ \hline
    {\bfseries III. Kalinga} & 881 & 111 & 110\\ \hline
    {\bfseries IV. Mughal} & 624 & 79 & 78\\
    \hline
    \end{tabular}
\end{table}

\begin{figure}
\begin{center}
\includegraphics[scale = 0.05]{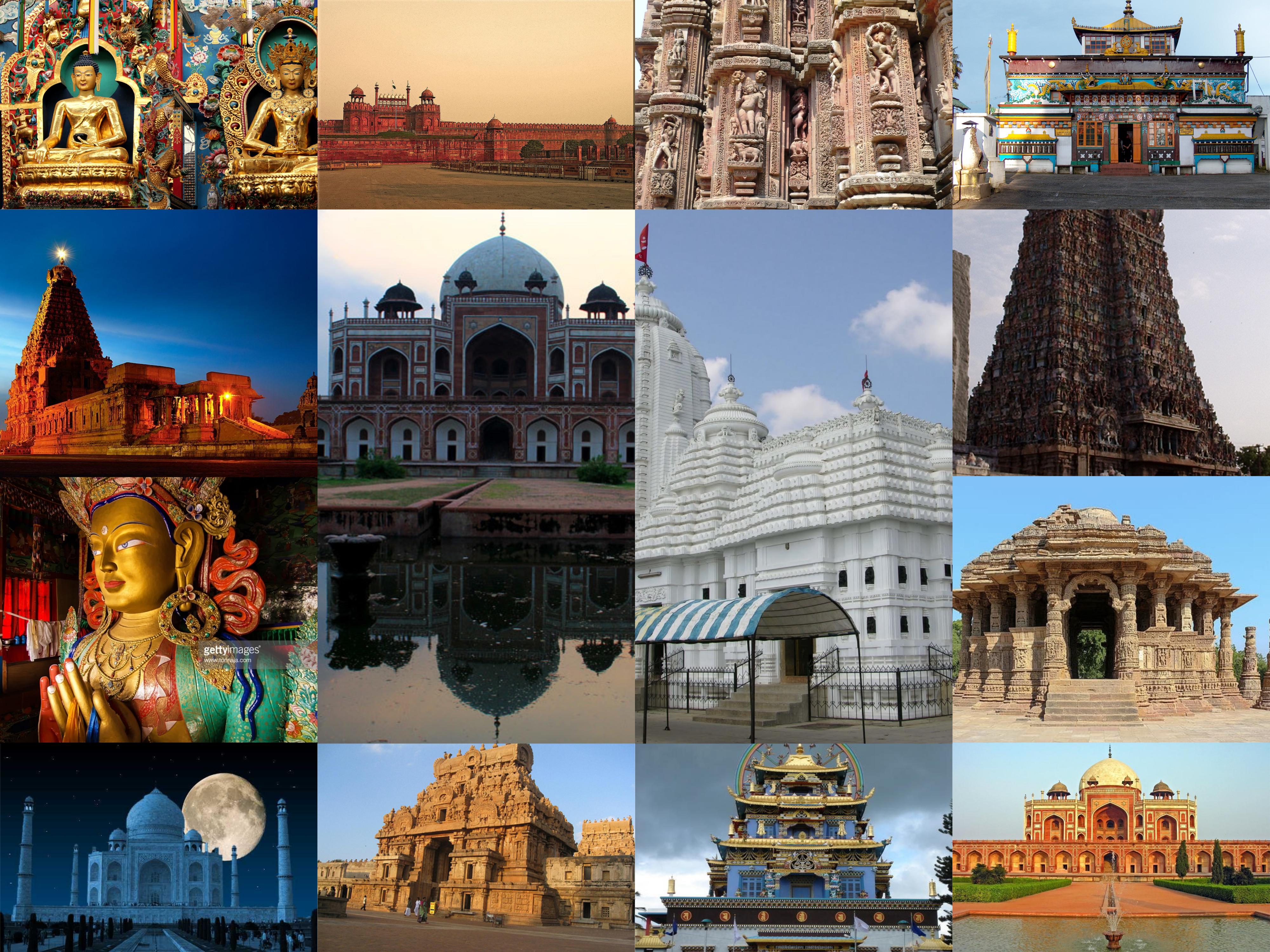}
\caption{Dataset Overview} \label{fig2}
\end{center}
\end{figure}

The total 3514 dataset images has been divided in ratio 80:10:10 of Training:Validation:Testing images respectively. Overview of the dataset is shown in the diagram below Fig. \ref{fig2}.
% \vspace{3pt}

% \vspace{3pt}

\section{Proposed Approaches}
In this section, we discuss about the proposed framework that is used for landmark classification. We devised two architectures to solve the problems of monument classification. These methods are described as follows: 

\subsection{Graph-based Visual Saliency (GBVS)}
\begin{figure}
\includegraphics[width=\textwidth]{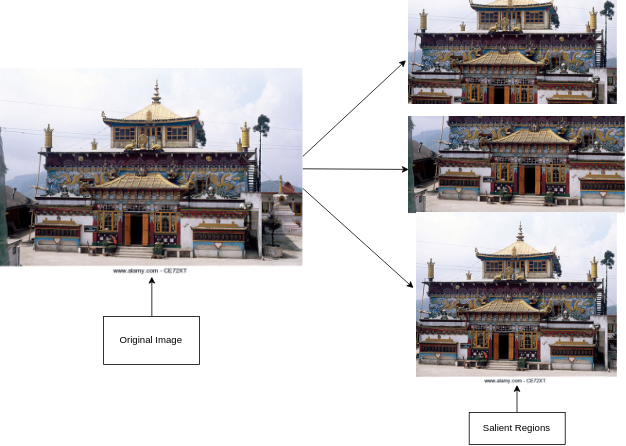}
\caption{Salient Region Detection using Graph-based Visual Saliency} \label{fig3}
\end{figure}

Image Saliency is what stands out and how fast you are able to quickly focus on the most relevant parts of what you see. Now, in the case of landmarks the less salient region is common backgrounds, that's of blue sky. The architectural design of the monuments is what differentiates between the classes. GBVS\cite{ref_11} \cite{ref_13} firstly finds feature maps and then apply non-linear Activation maps to highlight "significant" locations in the image. We used GBVS to detect 5 important locations per training image. Those images were used for multi-stage training. It helped to improve our accuracy by 3-4\%. Example of salient region detected using GBVS algorithm is shown in Fig. \ref{fig3}.

\subsection{Supervised Feature Classification}
In this approach, we used \textit{fc} layer features of ImageNet models to train supervised machine learning models such as kNN and Random Forest Classification. Among all the ImageNet models,  Inception ResNet V2 performed best for landmark classification. Therefore, we extracted the representation from Inception Net of dimension 2048 $\times$ 1.

\subsection{Ensemble Model Architecture}

\begin{figure}
\includegraphics[width=\textwidth]{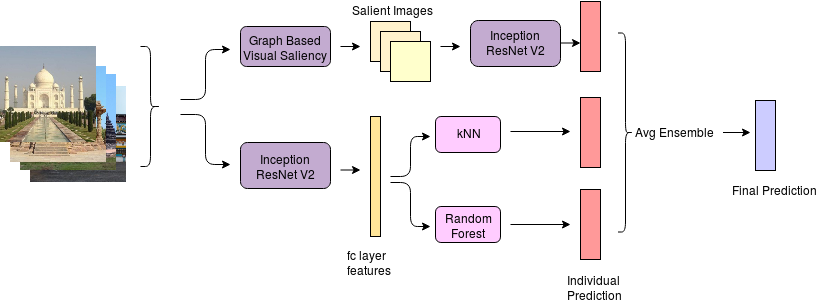}
\caption{Proposed Architecture} \label{fig4}
\end{figure}

Our final architecture comprises of Averaging based Ensemble \cite{ref_9} methods. Test image is passed firstly from GBVS algorithm to create a batch of 5 images. The batch prediction is done using Inception ResNet V2 \cite{ref_3}. Similarly, the test image is also passed through Inception ResNet model for feature extraction. These features were used to learn and predict classes using kNN and Random Forest Classifiers. The final prediction is done using averaging of predictions from the three models described above. Ensemble learning boosted the accuracy by approximately 2-3\%. The final architecture is diagramatically explained in Fig. \ref{fig4}.

\section{Experiments}
\subsubsection{Improved Feature Learning using Multi-stage Training}
We trained our model firstly on original images that were resized to 416 $\times$ 416 and then on high salient regions extracted using Graph-based Visual Saliency Algorithm. We used ImageNet pretrained weights to train Inception ResNet V2 \cite{ref_3} architecture on original and salient images. The salient images helped us to learn discriminative features between various classes. Original images assisted in learning of global spatial features. 

\subsubsection{Parameters} In our model, we used ADAM (\textit{lr}= 0.0001) \cite{ref_6} optimizer and ReLU\cite{ref_5} activation function. The model was trained for 7 epochs using the pretrained ImageNet weights as initialization for Transfer Learning. 

\section{Results}
The experimental results on the landmarks dataset are presented in Table \ref{tab2}. 
The scores obtained are from different architectures trained on salient crops and original images during Multi-stage Training.

\begin{table}
\centering
\caption{Accuracy during Multi-Stage Training on Inception V3 and Inception ResNet V2 models}\label{tab2}
\begin{tabular}{|l|l|l|l|l|}
\hline
\textbf{Model Architecture} &  \textbf{Data Subset} & \textbf{Train} & \textbf{Validation} & \textbf{Test}\\
\hline
Inception V3 \cite{ref_4} &  Original Images & 90.1 &77.23&75.42\\
\cline{2-5}
 &  Original + Salient & 91.81&80.3&78.91\\
 \cline{1-5}
Inception Resnet V2 \cite{ref_3} & Original Images & 91.76& 77&76.35\\
 \cline{2-5}
 & Original + Salient & 92.29 &81 &80\\

\hline
\end{tabular}
\end{table}

Table-\ref{tab3} compares the accuracy scores for all the models on train, validation and testing dataset. The final prediction is done by average ensembling of three models to get the final architecture with low variance and low bias. 

\begin{table}
\centering
\caption{Evaluation comparison (in \%) of different models}\label{tab3}
\begin{tabular}{|l|l|l|l|}
\hline
\textbf{Model Architecture} &  \textbf{Train} & \textbf{Validation} & \textbf{Test}\\
\hline
GBVS + InceptionResNetV2 & 92.61&89.65 & 86.18 \\
\hline
InceptionResnetV2 + kNN & 93.62&90.72 & 86.94\\
\hline
InceptionResNetV2 + Random Forest &91.58 & 89.8&88\\
\hline
Average Ensemble &  \textbf{94.58} & \textbf{93.8}&\textbf{90.08}\\
\hline
\end{tabular}
\end{table}

Table \ref{tab4} compares the results on the existing dataset with our new proposed approach. It is clear that our approach the outperforms the existing models by 8\%.
\begin{table}[]
    \centering
    \caption{Comparison of our best model with competing methods\cite{ref_8}}
    \begin{tabular}{|l|l|}
    \hline
         \textbf{Framework} & \textbf{Test}\\
         \hline
         SIFT + BoVW & 51\% \\
         \hline
         Gabor Transform + Radon Barcode & 70\%\\
         \hline
         Radon Barcode & 75\%\\
         \hline
         CNN & 82\%\\
         \hline
         \textbf{Our Method (Average Ensemble)} & \textbf{90\%}\\
         \hline
    \end{tabular}
    \label{tab4}
\end{table}

\section{Conclusion and Future Work}

The paper presented two approaches on which extensive experiments were done to classify Indian architectural styles. Landmark Recognition problem presents some noteworthy solutions as there are no training data available for less popular landmarks. Our solution focusses on the most noticeable region of the image to classify landmarks accurately. Our approach targets the fine-grained features as well as on the global features of monuments. Previous works lack the attention mechanism to differentiate models on the basis of fine-grained features. Our model outperforms the existing approach by 8\%. 

In future, the authors aim to improve the model accuracy by using DELF features and Visual Attention mechanism to further improve the accuracy of the model as well as learn more substantial features.


\begin{thebibliography}{8}

\bibitem{sift}
D. G. Lowe, “Distinctive image features from scale-invariant keypoints,” International Journal of Computer Vision, vol. 60, no. 2, pp. 91–110, 2004.
\doi{10.1023/B:VISI.0000029664.99615.94}

\bibitem{ref_1}
P. Shukla, B. Rautela and A. Mittal, "A Computer Vision Framework for Automatic Description of Indian Monuments," 2017 13th International Conference on Signal-Image Technology \& Internet-Based Systems (SITIS), Jaipur, 2017, pp. 116-122.
\doi{10.1109/SITIS.2017.29}

\bibitem{ref_2}
Z. Chen, A. Jacobson, N. Sunderhauf, B. Upcroft, L. Liu, C. Shen, I. Reid, M. Milford, Deep Learning Features at Scale for Visual Place Recognition, 2017 IEEE International Conference on Robotics and Automation (ICRA), Singapore.
\doi{10.1109/ICRA.2017.7989366}

\bibitem{ref_3}
Szegedy, Christian et al. “Inception-v4, Inception-ResNet and the Impact of Residual Connections on Learning.” AAAI (2017).

\bibitem{ref_4}
C. Szegedy, V. Vanhoucke, S. Ioffe, J. Shlens and Z. Wojna, "Rethinking the Inception Architecture for Computer Vision," 2016 IEEE Conference on Computer Vision and Pattern Recognition (CVPR), Las Vegas, NV, United States, 2016, pp. 2818-2826.\doi{10.1109/CVPR.2016.308}

\bibitem{ref_5}
Vinod Nair and Geoffrey E. Hinton. “Rectified Linear Units Improve Restricted Boltzmann Machines”. In Proceedings of the 27th International Conference on International Conference on Machine Learning. ICML’10. Haifa, Israel: Omnipress, 2010, pp.807–814. ISBN : 978-1-60558-907-7. URL : http://dl.acm.org/citation.cfm?id=3104322.3104425.

\bibitem{ref_6}
Diederik P. Kingma and Jimmy Ba. “Adam: A Method for Stochastic Optimization”. In: CoRR abs/1412.6980 (2014). arXiv: 1412.6980. URL:http://arxiv.org/abs/1412.6980.


\bibitem{ref_7}
G. Amato and F. Falchi and P. Bolettieri, "Recognizing Landmarks Using Automated Classification Techniques: Evaluation of Various Visual Features,". 2010 Second International Conferences on Advances in Multimedia. 78-83 
\doi{10.1109/MMEDIA.2010.20}
 

\bibitem{ref_8}
Sharma S., Aggarwal P., Bhattacharyya A.N., Indu S. (2018) Classification of Indian Monuments into Architectural Styles. NCVPRIPG 2017. Communications in Computer and Information Science, vol 841. Springer, Singapore.
\doi{10.1007/978-981-13-0020-2\_47}

\bibitem{ref_9}
Sainin, Mohd \& Alfred, Rayner \& Adnan, Fairuz \& Ahmad, Faudziah. (2018). Combining Sampling and Ensemble Classifier for Multiclass Imbalance Data Learning. 262-272. 
\doi{10.1007/978-981-10-8276-4\_25.}

\bibitem{surf}
Bay, Herbert and Ess, Andreas and Tuytelaars, Tinne and Van Gool, Luc, "Speeded-Up Robust Features (SURF),". Comput. Vis. Image Underst. Journal. June, 2008. Vol. 110. 346--359.
\doi{10.1016/j.cviu.2007.09.014}

\bibitem{hog}
Dalal, Navneet and Triggs, Bill, "Histograms of Oriented Gradients for Human Detection,". Proceedings of the 2005 IEEE Computer Society Conference on Computer Vision and Pattern Recognition (CVPR'05) - Volume 1 - Volume 01. 886--893. 
\doi{10.1109/CVPR.2005.177}

\bibitem{ref_10}
A. Saini, T. Gupta, R. Kumar, A. K. Gupta, M. Panwar and A. Mittal, "Image based Indian monument recognition using convoluted neural networks," 2017 International Conference on Big Data, IoT and Data Science (BID), Pune, 2017, pp. 138-142.
\doi{10.1109/BID.2017.8336587}

\bibitem{ref_11}
TRIANTAFYLLIDIS, Georgios; KALLIATAKIS, Gregory. Image based Monument Recognition using Graph based Visual Saliency. ELCVIA Electronic Letters on Computer Vision and Image Analysis, [S.l.], v. 12, n. 2, p. 88-97, apr. 2013. ISSN 1577-5097. 
\doi{10.5565/rev/elcvia.524}

\bibitem{ref_12}
B. Ghildiyal, A. Singh and H. S. Bhadauria, "Image-based monument classification using bag-of-word architecture," 2017 3rd International Conference on Advances in Computing,Communication \& Automation (ICACCA) (Fall), Dehradun, 2017, pp. 1-5.
\doi{10.1109/ICACCAF.2017.8344728}

\bibitem{ref_url1}
Google AI Blog:Google-Landmarks: A New Dataset and Challenge for Landmark Recognition, \url{https://ai.googleblog.com/2018/03/google-landmarks-new-dataset-and.html}.

\bibitem{ref_13}
Harel, Jonathan and Koch, Christof and Perona, Pietro (2007) Graph-Based Visual Saliency. In: Advances in Neural Information Processing Systems 19 (NIPS 2006). Advances in Neural Information Processing Systems. No.19. MIT Press , Cambridge, MA, pp. 545-552. ISBN 0-262-19568-2.

\end{thebibliography}
\end{document}